\documentclass[aps,prx,reprint,amsmath,amssymb,groupedaddress,floatfix]{revtex4-2}

\usepackage{graphicx}
\graphicspath{{Figures/}{./}}
\usepackage[T1]{fontenc}
\usepackage[utf8]{inputenc}
\usepackage{float}
\usepackage{algorithmic}
\usepackage[colorlinks,linkcolor=blue,anchorcolor=blue,citecolor=blue,urlcolor=blue]{hyperref}
\usepackage{cleveref}
\newcounter{algorithm}
\newenvironment{algorithm}[1][]{%
  \begin{figure}[H]
  \small\raggedright
  \hrule\vspace{2pt}
  \def\caption##1{\refstepcounter{algorithm}\noindent\textbf{Algorithm~\thealgorithm}\ ##1\par\vspace{2pt}\hrule\vspace{3pt}}%
}{%
  \vspace{2pt}\hrule
  \end{figure}
}
\crefname{algorithm}{Algorithm}{Algorithms}

\begin{document}

\title{Network dynamics-based framework for understanding deep neural networks}

\author{Yuchen Lin$^{1}$}
\author{Sihan Feng$^{1,3}$}
\author{Yong Zhang$^{1,2}$}
\thanks{Contact author: yzhang75@xmu.edu.cn}
\author{Hong Zhao$^{1,2}$}
\thanks{Contact author: zhaoh@xmu.edu.cn}
\affiliation{\textsuperscript{1}Department of Physics, Xiamen University, Xiamen 361005, China\\
\textsuperscript{2}Lanzhou Center for Theoretical Physics, Lanzhou University, Lanzhou 730000, China\\
\textsuperscript{3}Shenzhen Angel FOF Management Co., Ltd., Shenzhen 518000, China}

\date{\today}

\begin{abstract}
Advancements in artificial intelligence demand a deeper understanding of the underlying mechanisms of deep learning. This study, based on the theory of nonlinear dynamical systems, constructs a theoretical framework for analyzing deep networks at a microscopic level. The framework consists of two main components: (1) a classification of information transformation modes. We categorize the ways in which a network transforms information layer by layer into two basic types: order-preserving transformations and non-order-preserving transformations. Order-preserving transformations can be achieved by individual neurons, while non-order-preserving transformations require the cooperation of multiple neurons. (2) A dynamical method for performance evaluation. By introducing the concept of attraction basins in the sample and weight spaces, we characterize the network's performance from the perspective of dynamical system stability. The attraction basins in the sample space reflect the generalization ability of the network, while the attraction basins in the weight space represent structural robustness. Different information transformation modes lead to different distributions of weight vectors in space. By identifying these structures, we can estimate the relative contributions of each transformation mode in different layers, thus revealing distinct learning ``phases''. Based on this, we explain a key theoretical source of deep network performance advantages, provide a mechanistic explanation for the phenomenon of ``grokking'', establish the theoretical basis for the existence of an optimal depth, and offer theoretical guidance for selecting hyperparameters such as learning rate and batch size.
\end{abstract}

\maketitle

\section{Introduction}\label{section1}
To understand deep neural networks (DNNs), several influential theoretical frameworks have been developed, including the information bottleneck theory \cite{7133169,alemi2017deep,PhysRevLett.132.197201}, flatness-based landscape analysis \cite{doi:10.1073/pnas.1608103113,ma2021on,doi:10.1073/pnas.2015617118,PhysRevLett.130.237101,10.1162/neco.1997.9.1.1,keskar2017on, feng2023activity}, geometric approaches \cite{JMLR:v21:20-345, Xie2025LCL, Cheng2021PRB}, group-theoretic methods \cite{doi:10.1073/pnas.2016917118}, as well as model linearization analysis \cite{jacot2018neural, arora2019exact,saxe2014a,ji2018gradient} and shallow network approximations \cite{dandi2024two,JMLR:v22:20-1123, Han2018PRX}.  Theoretical efforts have also expanded to explain key empirical phenomena in DNNs, such as the  ``double descent''  phenomenon \cite{doi:10.1073/pnas.1903070116,schaeffer2024double,Nakkiran2020Deep}, grokking \cite{power2022grokkinggeneralizationoverfittingsmall,kumar2024grokking,fan2024deepgrokkingdeepneural, hu2025llmslearnreasoncomplex, tian2025provablescalinglawsfeature, Guo2025}, discontinuous learning mechanisms \cite{doi:10.1073/pnas.2215352119,Huang2024SCPMA}, neural scaling laws \cite{kaplan2020scaling,doi:10.1073/pnas.2311878121}, among others. Although these contributions have significantly advanced our understanding of machine learning mechanisms, most focus on the global behavior of the network as a whole and have yet to effectively bridge local structural properties with overall performance. As a result, deep neural networks continue to be widely regarded as a ``black box'' \cite{Zhao2021BlackBox, Chen2025AcceleratorML, Zhou2022SCPMA, Zhang2026PINNs, He2023NuclearML}.

Despite neurons' foundational role as computational units with linear summation and nonlinear activation, existing theoretical approaches have yet to fully establish how neuron-level properties impact global learning dynamics. This limitation is evident in the difficulty of explicitly defining the nonlinearity of learning models at the neuron level and integrating these basic building blocks to explain local inter-neuron interactions leading to system-level learning phenomena. The conventional linear/nonlinear classification of learning models based on activation functions lacks precision, as demonstrated by the fact that networks with nonlinear activation functions often initially exhibit linear-like dynamics before gradually developing their full nonlinear characteristics during training \cite{GEIGER20211,xu2024mechanismsfeaturelearningexact,Geiger_2020}.

\begin{figure}[ht]
\centering
\includegraphics[width=1\linewidth]{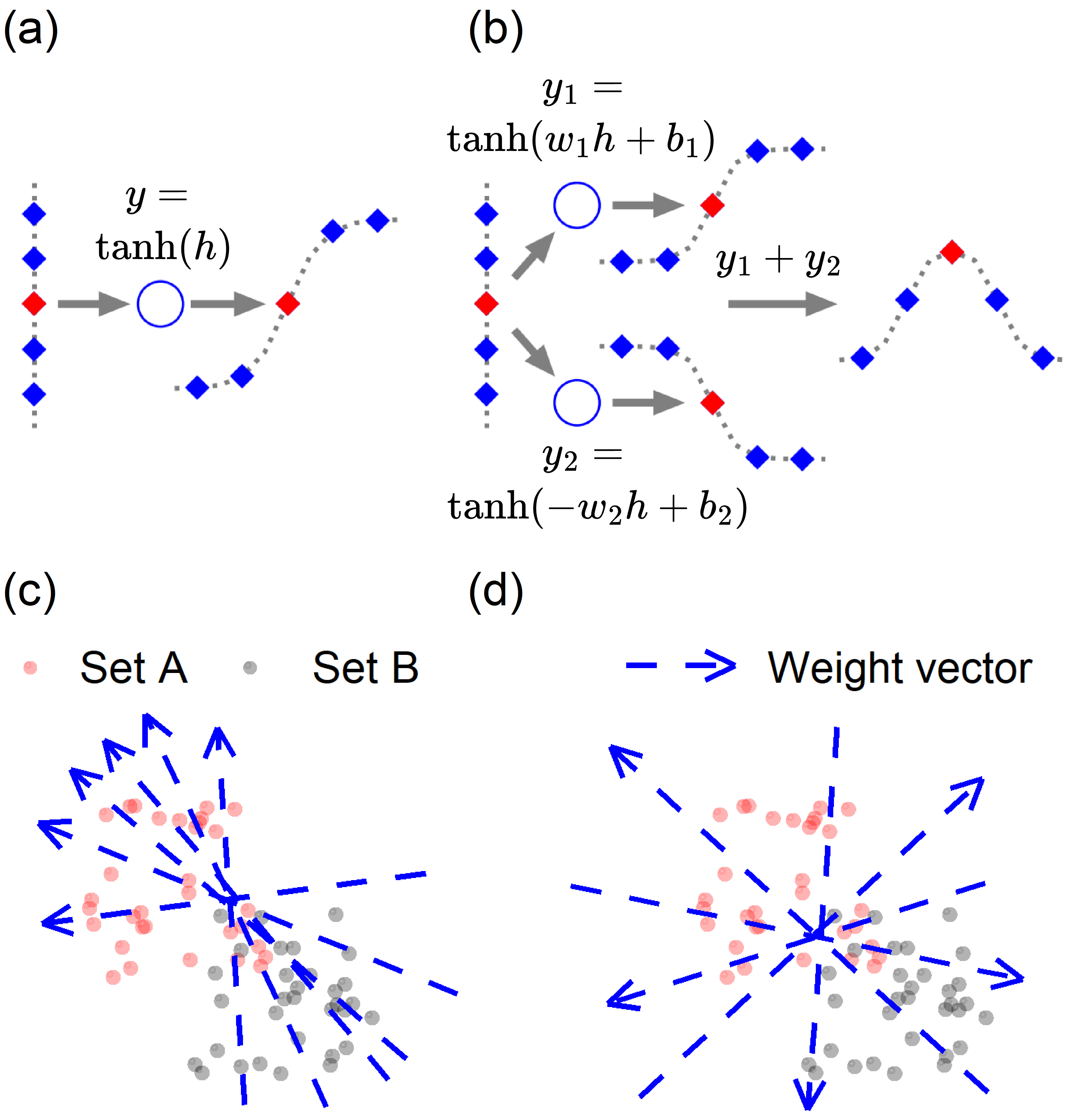}
\caption{Illustration of transformation modes and their effects. The solid diamonds represent the local fields of five samples projected by a weight vector; hollow circles represent neurons. (a) The OPT mode preserves the order of the samples' local fields and can be achieved by a single neuron. (b) The NPT mode alters the order, typically requiring at least two cooperating neurons (under a typical monotonic nonlinear activation function). (c) The weight vectors induced by OPT are concentrated to maximize the output for sample set A. (d) The weight vectors induced by NPT are isotropic, as they do not rely on projection maximization to extract information.}
\label{fig1}
\end{figure}

This paper introduces a neuron-level analytical framework for investigating learning dynamics, which is founded upon two novel concepts. First, we introduce the concept of fundamental transformation units, operating at the individual neuron level. For a training set of \( P \) samples, each neuron processes \( P \) local fields as an input sequence. This sequence undergoes transformation via two distinct modes: order-preserving transformation (OPT), where the input sequence's ordering is maintained (Fig.~\ref{fig1}(a)), and Non-order-preserving transformation (NPT), where the input sequence's ordering is altered (Fig.~\ref{fig1}(b)). The ordering denotes the ranking of local fields from largest to smallest. Crucially, OPT operations are effectively linear in terms of order preservation and can be implemented by most monotonic activation functions (including linear activation as a special case). In contrast, NPT operations, typically achieved through non-monotonic activation functions (e.g., Gaussian) or specific combinations of common activation functions like ReLU or tanh, produce localized peak-shaped responses via nonlinear folding operations. These distinct transformation modes profoundly influence the distribution of weight vector directions and thus information extraction (Figs.~\ref{fig1}(c) and (d)). Consequently, the OPT/NPT ratio emerges as a quantitative measure of nonlinearity, offering an interpretable design parameter for optimizing information processing.

Second, we introduce the concept of attraction basins in the sample space and the weight space. The sample-space basin captures input-output sensitivity, while the weight-space basin reflects stability in the parameter space. These two attraction basins influence each other, and their balance provides the selection criteria for deep neural network architectural parameters (such as depth and width) and training strategies (such as learning rate and batch size). The two attraction basins complement the flat minima analysis. The latter establishes a dual relationship between the sensitivity of the loss function to weight perturbations and sample perturbations, akin to perturbation analysis. In contrast, we investigate the boundaries at which a trained network resists sample perturbations and weight perturbations, revealing that the two attraction basins can vary independently. 

Our proposed transformation modes describe the local operations of the learning process, while the attraction basins characterize its emergent behavior. We can analyze the transformation modes used during the learning process at multiple levels—layer by layer, sample by sample, and neuron by neuron. Additionally, we leverage multi-level attraction basins, including overall averages, class-specific basins, and sample-specific basins. In this way, we provide a framework that illuminates every local aspect of the so-called ``black box'' of DNNs and connects these aspects to overall performance. Moreover, this perspective allows us to view DNNs as layer-wise iterative dynamical mappings, offering a powerful lens grounded in dynamical systems theory to better understand their learning behavior.

A key feature revealed by this framework is the emergence of distinct learning phases during training, each characterized by specific OPT/NPT state distributions across layers. These phases are closely linked to network performance. As an illustrative example, we conduct a detailed case analysis of the well-known grokking phenomenon, which is marked by sudden improvements in test accuracy after prolonged periods of near-random performance, thereby illustrating how phase transitions align with qualitative shifts in model performance. Although both exhibit phase transition behavior, attraction basin analysis reveals that grokking in deep networks and in the reference shallow networks \cite{liu2023omnigrok} occurs via distinct mechanisms.

The paper introduces transformation modes and attraction basins, along with methods for their quantitative characterization. It then analyzes learning dynamics in shallow models, emphasizing phase emergence and transitions, and further examines deep neural networks, focusing on how architecture and training parameters shape key dynamical metrics. Finally, it explores the mechanisms underlying phase transitions in grokking and concludes with broader implications and future directions.

\section{Basic concepts}\label{sec:2}
\subsection{Model} 
Without loss of generality, we consider a fully connected DNN architecture for classification, defined by the following equations:

\begin{equation}
  \begin{aligned}
   x^{(l)}_{i_l} &= f(h^{(l)}_{i_l}),\\
   h^{(l)}_{i_l} &= \sum_{i_{l-1}=1}^{N_{l-1}} w^{(l)}_{i_l i_{l-1}} x^{(l-1)}_{i_{l-1}},
  \end{aligned}
\label{eq:eq1}
\end{equation}

where \( x^{(l)}_{i_l} \) denotes the output of the \( i_l \)-th neuron in layer \( l \), and \( h^{(l)}_{i_l} \) is the corresponding local field. The weight \( w^{(l)}_{i_l i_{l-1}} \) connects the \( i_{l-1} \)-th neuron in layer \( l-1 \) to the \( i_l \)-th neuron in layer \( l \). The function \( f(\cdot) \) represents the activation function, and \( N_{l-1} \) denotes the number of neurons in layer \( l-1 \). Note that we label the input layer as \( l = 1 \); accordingly, weights \( w^{(l)} \) are defined for \( l = 2, \dots, L \). For notational uniformity in the recursive form, we formally introduce \( x^{(0)} \equiv x^{(1)} \) when applying \cref{eq:eq1}. For simplicity of theoretical analysis, bias terms are omitted, as they do not affect the structural properties considered here and can be incorporated straightforwardly if needed. All computations are implemented in PyTorch, a widely used deep learning framework.

\subsection{Fundamental transformation units of a learning model}

Our approach is grounded in a fundamental postulate: classification depends on the mutual information between samples. Specifically, a weight vector \( \mathbf{w} \)—a row of the weight matrix \( \mathbf{W} \)—projects the \(P\) sample vectors \( \mathbf{x}^{\mu} \) onto a sequence \( h(\mu) \) with \( \mu = 1, 2, \ldots, P \), where \(h(\mu) = \mathbf{w} \cdot \mathbf{x}^{\mu}.\)
It is not the absolute value of \( h(\mu) \) itself, but rather its relational properties—namely, its position within the sequence and its ordering relative to the projections of other samples—that define the features distinguishing this sample from the rest.

The objective of training is twofold: first, to maximize such mutual information through the choice of appropriate weight vectors; second, to employ the neuron’s transfer function to convert this mutual information into a form that can be effectively utilized by the corresponding output-layer neuron.

The neuron’s transfer function operates in only two distinct modes: OPT and NPT.

\textbf{(1) The OPT mode} preserves the ordering of the input sequence \( h(\mu) \), as illustrated in Fig.~\ref{fig1}(a). This behavior can be realized even with strongly nonlinear but monotonic activation functions such as ReLU or \(\tanh\). The transformation is effectively linear in terms of preserving input ordering, with linear activation representing a special case of this mode. In this mode, for a sample to yield the largest output among all samples, its local field must attain the maximum value within the projected sequence of local fields.

\textbf{(2) The NPT mode} alters the ordering of the input sequence. This can arise either from an individual neuron with a non-monotonic activation, such as the Gaussian-type function \( y = \exp(-h^2) \), or from combinations of neurons. For monotonic activations like \(\tanh\), the minimal combination of two neurons, e.g., \( y = \tanh(w_1 h + b_1) + \tanh(-w_2 h + b_2) \), is sufficient to produce a maximum at an arbitrary position within the input sequence. For clarity, we assume identical input sequences across the two neurons. ReLU activations exhibit a similar capability. Both cases can result in folding-like transformations, as illustrated in Fig.~\ref{fig1}(b). More complex combinations of neurons can further enhance this effect. This mode is inherently nonlinear in its functional consequences, consistent with the notion of nonlinearity in dynamical systems, where the emergence of intrinsic nonlinear behavior of chaos requires both stretching and folding, rather than merely the presence of nonlinear terms in the governing equations.

Thus, we define fundamental local processing units that perform qualitatively distinct transformations—either linear (OPT) or nonlinear (NPT). These modes give rise to different collective behaviors of the weight vectors. In the OPT mode, weight vectors must align with specific directions to promote the local fields of target samples to higher ranks within the projected sequence, thereby maximizing output activations, as illustrated in Fig.~\ref{fig1}(c). This behavior is effective for extracting linearly separable features. However, in order to minimize the training loss for rare or atypical samples, certain weight vectors may converge to specific orientations that maximize specific features of those samples. While this may reduce loss locally, it risks overfitting and compromises generalization. Moreover, this mode cannot distinguish linearly inseparable samples.

In contrast, the NPT mode imposes no directional constraints on the weight vectors, enabling information extraction from a broader set of orientations, as illustrated in Fig.~\ref{fig1}(d). This flexibility allows neurons to capture more complex input relationships.  From the perspective of information extraction, the weight vector distribution in Fig.~\ref{fig1}(d) is superior to that in Fig.~\ref{fig1}(c). However, operating in a more nonlinear regime also makes neurons more sensitive to input variations. An effective learning model should thus integrate neurons operating in both modes to balance expressiveness and stability.

We note that the related concepts of OPT and NPT modes were introduced in our earlier preliminary work \cite{feng2022learnthemodesmachinelearning}, where we incorrectly assumed that the former extracts linearly separable information while the latter extracts linearly inseparable information. In fact, even on linearly separable datasets (such as MNIST), NPT plays an indispensable role in boosting test accuracy, as the present study seeks to uncover.

\subsection{Rank probability distribution (RPD) and linear substitution map (L-Map)}
The core question that remains is how to characterize the distribution of OPT and NPT neurons in each hidden layer and thus quantify the linearity layer by layer. We propose the following metrics to address these issues, first introducing them in the case of a three-layer network with a single hidden layer, and then extending the analysis to DNNs.

Feeding all $P$ samples into the network yields, for each hidden-layer neuron, a sequence of local fields $h_i^{(2)}(\mu)$, where $\mu = 1, 2, \dots, P$. For the $i$-th neuron, we construct a signed projection sequence $h_i^{(2)}(\mu) \cdot \mathrm{sign}(W^{(3)}_{ki})$, associated with the connection to the $k$-th output neuron (corresponding to class $k$). For samples belonging to class $k$, the output neuron corresponding to class $k$ is expected to produce higher activations. Under the OPT mode, the ordering of the transformed sequence is preserved, and neurons operate independently. Thus, to maximize the output activation, the values $h_i^{(2)}(\mu) \cdot \mathrm{sign}(W^{(3)}_{ki})$ for class $k$ samples should achieve higher ranks within the sequence, resulting in a collective alignment of weight vectors. In contrast, under the NPT mode, ordering is not preserved, and maximization can be achieved through combinations of neurons without requiring collective alignment.

To characterize the above effects, we rank the samples belonging to class $k$ among all $P$ samples according to their values in the projection sequence. Without loss of generality, we sort the sequence in descending order, i.e., samples with larger $h_i^{(2)}(\mu)\cdot \mathrm{sign}(W^{(3)}_{ki})$ values have higher ranks. The ranks occupied by samples belonging to class $k$ then form a set of rank positions, whose empirical distribution defines the RPD for the given neuron and class. Repeating this procedure for all output classes yields, for each hidden neuron, a collection of rank positions characterizing its overall contribution to classification. Applying the same methodology, we obtain RPDs for all neurons in the hidden layer. Since our primary interest lies in statistical behavior, we pool the rank statistics over all classes and neurons, yielding a smooth RPD that characterizes the overall property of the layer. When needed, class-resolved RPDs can be examined to diagnose category-specific deviations from the layer-averaged behavior. Importantly, the RPD is computed using the pre-activation local fields $h_i^{(2)}$; since commonly used activation functions are monotonic, they do not alter the relative ordering of samples and therefore do not affect the rank statistics. The RPD captures the collective alignment properties of weight vectors, as illustrated in Figs.~\ref{fig1}(c) and (d). In other words, the steepness of the RPD provides a quantitative probe of the relative proportions of OPT and NPT neurons in a given hidden layer.

To extend the analysis to DNNs, we introduce the L-map as follows. Specifically, to characterize the ranking behavior of the $l$-th hidden layer, we first introduce the corresponding $\mathbf{W}^{(l)}_{\mathrm{L\text{-}map}}$, which encodes the effective linear mapping from the $l$-th layer to the output layer. By replacing all nonlinear activation functions beyond the $l$-th layer with the identity function $f(h)=h$, we obtain

\begin{equation}
\mathbf{W}^{(l)}_{\mathrm{L\text{-}map}} = \mathbf{W}^{(L)} \cdot \mathbf{W}^{(L-1)} \cdots \mathbf{W}^{(l+1)},
\label{eq:eq2}
\end{equation}

where $\mathbf{W}^{(l)}$ denotes the weight matrix of the $l$-th layer. We then define the output of the L-map as:
\[
\mathbf{H}^{(l)}_{L}=\mathbf{W}^{(l)}_{\mathrm{L\text{-}map}}\cdot \mathbf{X}^{(l)},
\]
where \(\mathbf{X}^{(l)} \in \mathbb{R}^{N_l \times P}\) denotes the post-activation output of the \(l\)-th layer, with each column corresponding to one sample. The resulting $\mathbf{H}^{(l)}_{L}$ thus represents the network output after applying the L-map. The matrix $\mathbf{W}^{(l)}_{\mathrm{L\text{-}map}}$ therefore serves as an effective output-layer weight matrix associated with the $l$-th layer.

Using $\mathbf{W}^{(l)}_{\mathrm{L\text{-}map}}$ together with the pre-activation local fields $\mathbf{H}^{(l)}$ of the $l$-th layer, the RPD of that layer can be computed following the same procedure as in the shallow-network case. With $\mathbf{W}^{(l)}_{\mathrm{L\text{-}map}}$ defined above, the L-map provides a layer-wise reduction of a DNN to an effective linear readout from layer $l$ to the output. If the subnetwork beyond layer $l$ is entirely linear, this reduction is exact: the L-map is mathematically equivalent to a linear perceptron and can substitute for the latter part of the network. Accordingly, in the RPD computation the output-layer connection $W^{(3)}_{ki}$ in the three-layer case is replaced by $(W^{(l)}_{\mathrm{L\text{-}map}})_{ki}$.

In DNNs with nonlinear activation functions, the same replacement is used as an approximation to estimate the proportion of neurons operating in the two modes. This constitutes an approximation, as nonlinear effects from subsequent hidden layers may affect the accuracy of the estimation. However, when a monotonic activation function such as tanh or ReLU is used, the accuracy of this approximation remains relatively high. Since the RPD depends only on the ranking of sample projections, preserving the sign of $(W^{(l)}_{\mathrm{L\text{-}map}})_{ki}$, rather than its absolute value, is sufficient. Replacing monotonic activation functions with linear ones maintains the input-output monotonicity, thereby enhancing the robustness of the sign preservation. Details of the procedure to calculate the RPD are provided in Appendix.

In practice, however, additional components such as batch normalization (BN) \cite{10.5555/3045118.3045167} are commonly employed during training to stabilize learning and accelerate convergence. BN standardizes activations within each mini-batch and applies a feature-wise affine transformation. These operations modify the network’s L-map and must be properly accounted for in the analysis.

BN operates differently in training and evaluation modes \cite{NEURIPS2019_bdbca288}. During training, it normalizes each layer using the mean and variance computed from the current mini-batch. In evaluation mode, it instead applies a moving average of the mean and variance accumulated throughout training. The updates follow the formulas:
\begin{equation}
  \begin{aligned}
    \hat{\mu}_t &= (1 - \alpha) \hat{\mu}_{t-1} + \alpha \mu_t, \\
    \hat{\sigma}^2_t &= (1 - \alpha) \hat{\sigma}^2_{t-1} + \alpha \sigma_t^2,
  \end{aligned}
  \label{eq:eq3}
\end{equation}
where \({\mu}_t\) and \({\sigma}^2_t\) are the mean and variance of the current mini-batch, and \(\hat{\mu}_t\), \(\hat{\sigma}^2_t\) are the accumulated estimates used in evaluation. The momentum parameter is typically set as \(\alpha = 0.1\). Setting \(\alpha = 1\) effectively corresponds to using only the current batch statistics, i.e., the behavior during training.

After normalization, BN applies a learnable scaling factor \( \gamma_i^{(l)} \) to each feature \( i \) in layer \( l \), enabling the network to recover suitable activation magnitudes. The combined normalization and rescaling can be expressed as a diagonal matrix \( \mathbf{D}^{(l)} = \text{diag} ( \gamma_i^{(l)} / \hat{\sigma}_i^{(l)} ) \), where \( \hat{\sigma}_i^{(l)} \) is the estimated standard deviation for feature \( i \). This matrix captures the feature-wise transformation introduced by BN at layer \( l \) during inference.

Consequently, the L-map matrix is updated to

\begin{equation}
\begin{aligned}
\mathbf{W}^{(l)}_{\mathrm{L\text{-}map}}
={}&(\mathbf{D}^{(L)}\mathbf{W}^{(L)})
(\mathbf{D}^{(L-1)}\mathbf{W}^{(L-1)}) \\
&\cdots(\mathbf{D}^{(l+1)}\mathbf{W}^{(l+1)}).
\end{aligned}
\label{eq:eq4}
\end{equation}

\subsection{Attraction basins}
Our second key analytical tool is the concept of attraction basins, a fundamental notion in nonlinear dynamical systems. The analysis of attraction basins has been applied to study the dynamics of asymmetric Hopfield neural networks \cite{PhysRevE.70.066137, 6795618, PhysRevE.72.066111}, where a sharp transition from a chaotic phase to a memory phase emerges as the basins expand \cite{6795618, PhysRevE.72.066111}. 
We extend the concept of attraction basins to the context of DNNs. Here, we define the attraction basin of a training sample in two distinct spaces: the sample space and the weight space.

One type of attraction basin is defined by applying random perturbations to a training sample and evaluating whether the model retains its original prediction. Specifically, if the trained network still classifies a perturbed version \( \mathbf{x}^\mu + \delta \mathbf{x} \) of the \( \mu \)-th sample into the same class, then \( \delta \mathbf{x} \) is considered to lie within the attraction basin of that sample. Each original input \( \mathbf{x}^\mu \) is first normalized to the range \([0, 1]\). Gaussian noise with a specified standard deviation is added, and the resulting sample is then rescaled to the original data range. By plotting the classification accuracy averaged over multiple perturbation trials at each noise amplitude, we observe a gradual decline from noise-free accuracy to the level expected from random guessing. Without loss of generality, we define the noise amplitude at which the accuracy falls to 50\% as the size of the sample's attraction basin in the sample space. This metric directly relates to the model's robustness to input variations.

Another type of attraction basin is defined by perturbing the network weights and assessing whether a training sample remains correctly classified. Specifically, if \( \mathbf{x}^\mu \) is still recognized as belonging to the same class under a perturbed weight configuration \( \mathbf{w} + \delta \mathbf{w} \), then \( \delta \mathbf{w} \) is regarded as lying within the attraction basin in weight space for that sample. For consistency, weights are mean–variance normalized, Gaussian noise of controlled magnitude is added, and the perturbed weights are inverse-transformed back to their original scale. The updated weights are used to evaluate classification accuracy, with the basin size defined, analogously to the sample space case, at the noise magnitude where accuracy falls to $50\%$. This type of attraction basin characterizes the network’s structural stability and its robustness to perturbations in the weight space.

\section{Applications to shallow networks}\label{sec:3}
In this section, we demonstrate how the above concepts can be applied to analyze a shallow three-layer neural network with the architecture \(784\text{–}2048\text{–}10\), trained on the MNIST dataset \cite{6296535}, a widely used benchmark for handwritten digit recognition consisting of 60000 training and 10000 test samples, each represented as a \(28 \times 28\) grayscale image.

\begin{figure}[ht]
\centering
\includegraphics[width=1\linewidth]{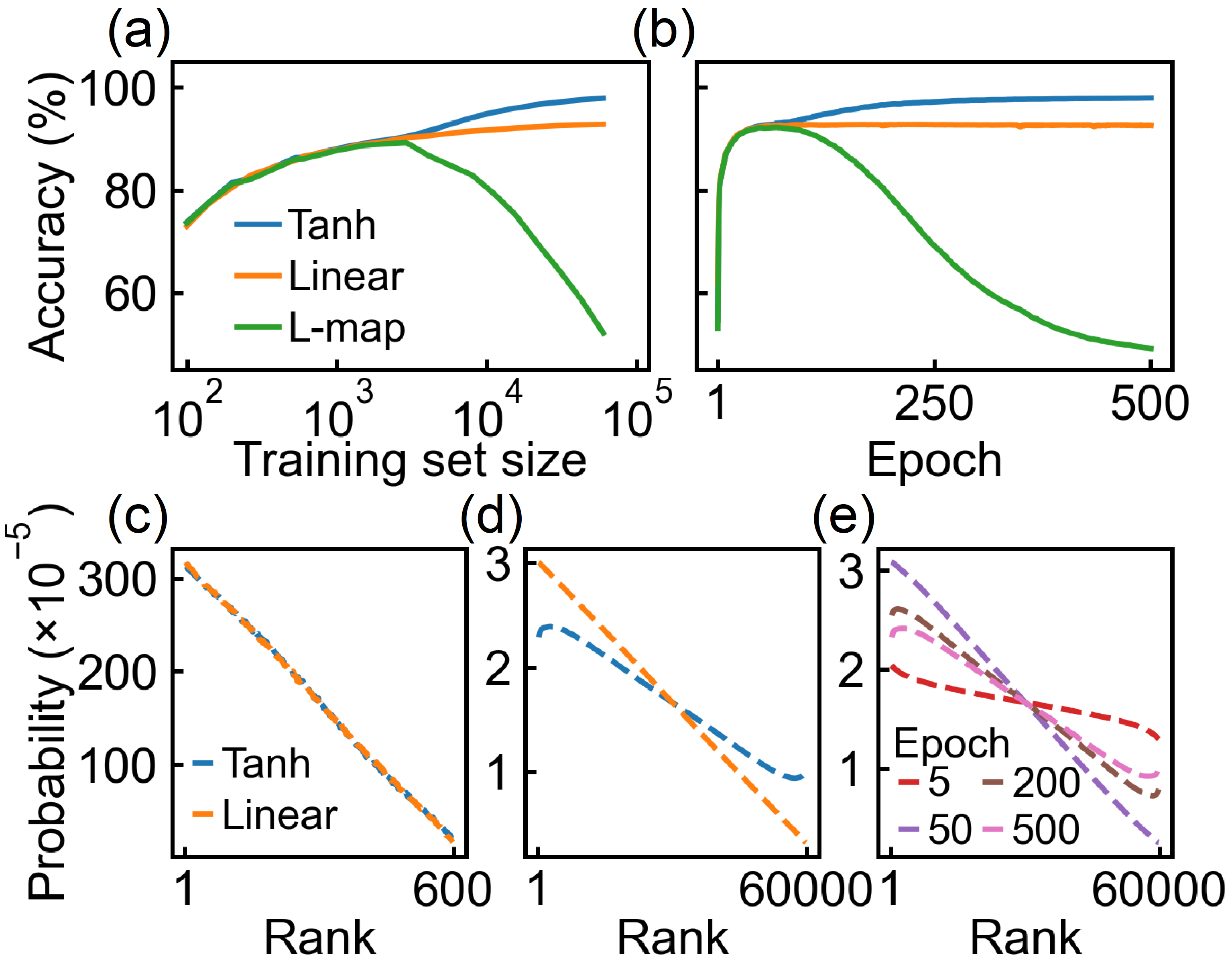}
\caption{Learning dynamics of a shallow network.  
(a) Test accuracy versus number of training samples for three models: the tanh network, an LNN with \(f(h)=h\), and the L-map, defined by replacing the activation function of the trained tanh network with the identity function while keeping the same parameters.
(b) Test accuracy versus training epochs using the full training set. Curve ordering matches (a). 
(c), (d) RPDs of the tanh network and the LNN after training on 600 and 60000 samples, respectively.
(e) RPDs of the tanh network at selected training epochs, showing their evolution over time.}
\label{fig2}
\end{figure}

Fig.~\ref{fig2}(a) shows the test accuracy as a function of training set size for three cases: the network with tanh activation, the linear neural network (LNN) with activation $f(h)=h$ trained independently from scratch, and the same tanh nonlinear network with identical trained weights but with the activation functions replaced by $f(h)=h$ without further training (the L-map). For small training sets, the accuracy curves coincide, whereas with increasing sample size, they begin to diverge at the same critical size.

Fig.~\ref{fig2}(b) shows the evolution of these test accuracies as a function of training epochs when the entire training set is used. In the early stage, all three curves overlap, and they start to diverge simultaneously as training proceeds.

This comparison shows that nonlinear networks effectively behave as linear ones when trained on small sample sets or during the early stage of training. Nevertheless, it remains unclear why nonlinear networks mimic linear behavior under these conditions.

Fig.~\ref{fig2}(c) presents the RPDs of the hidden layer for the nonlinear network and the LNN, trained on 600 samples. The RPDs show high density in high-ranking and low density in low-ranking regions, demonstrating the OPT-induced alignment of weight vectors. The close match between the two RPD curves suggests that learning is driven purely by OPT neurons, since the LNN can perform only the OPT operation.

Fig.~\ref{fig2}(d) shows that with the full training set, the two RPDs differ substantially, with the LNN exhibiting a steeper gradient. These observations
indicate that a significant number of NPT-mode neurons
are activated in the nonlinear network.

Although the nonlinear network ultimately deviates from the LNN, its early-time dynamics are effectively linear (Fig.~\ref{fig2}(b)). Consistent with this, the RPD gradient rises from an initially flat profile—set by isotropic random weights—to a peak as OPT-mode neurons align weight vectors along preferred directions, and then declines as training proceeds (Fig.~\ref{fig2}(e)). This alignment renders the nonlinear network LNN-like at early times; the subsequent decline reflects the recruitment of NPT-mode neurons, which disperses weight directions and reduces the RPD gradient. RPD analysis can also reveal the proportion of learning modes for each class and how they evolve during training, providing more detailed insights into the learning dynamics. Detailed results are presented in the Supporting Information.

These observations point to distinct learning phases. An initial OPT-dominated phase governs early learning. On small, linearly separable datasets, OPT alone often suffices to minimize the loss, and the system remains in this phase. For larger or more complex datasets, OPT becomes insufficient, triggering NPT activation and a transition to a mixed phase in which both OPT- and NPT-mode neurons contribute. 

These results further clarify the role of the L-map. When the substituted layer exhibits NPT modes, the L-map lowers the test accuracy because it disrupts these modes. In contrast, when the substituted layer operates in OPT modes, this substitution does not affect the accuracy. Together with the observation that the critical point at which the accuracy curves of the nonlinear and linear networks begin to separate coincides with the point where the L-map accuracy curve diverges, this indicates that the L-map provides a method to quantify the degree of linearization in a nonlinear network without the need to construct a separate LNN.

\section{Applications to DNNs}\label{sec:4}
In this section, we adopt the ReLU activation function for nonlinear DNNs and the identity function \( f(h) = h \) for linear DNNs. All models have an input dimension of 784 and an output dimension of 10, and are trained on the full MNIST training set. This investigation aims to reveal how optimization algorithms (SGD vs.\ Adam), hyperparameters (learning rate and batch size), and network architectures (depth and width) influence the layer-wise RPD distribution and the evolution of both attraction basins, thereby elucidating their underlying mechanisms.

\subsection{RPD analysis}
\begin{figure}[ht]
\centering
\includegraphics[width=1\linewidth]{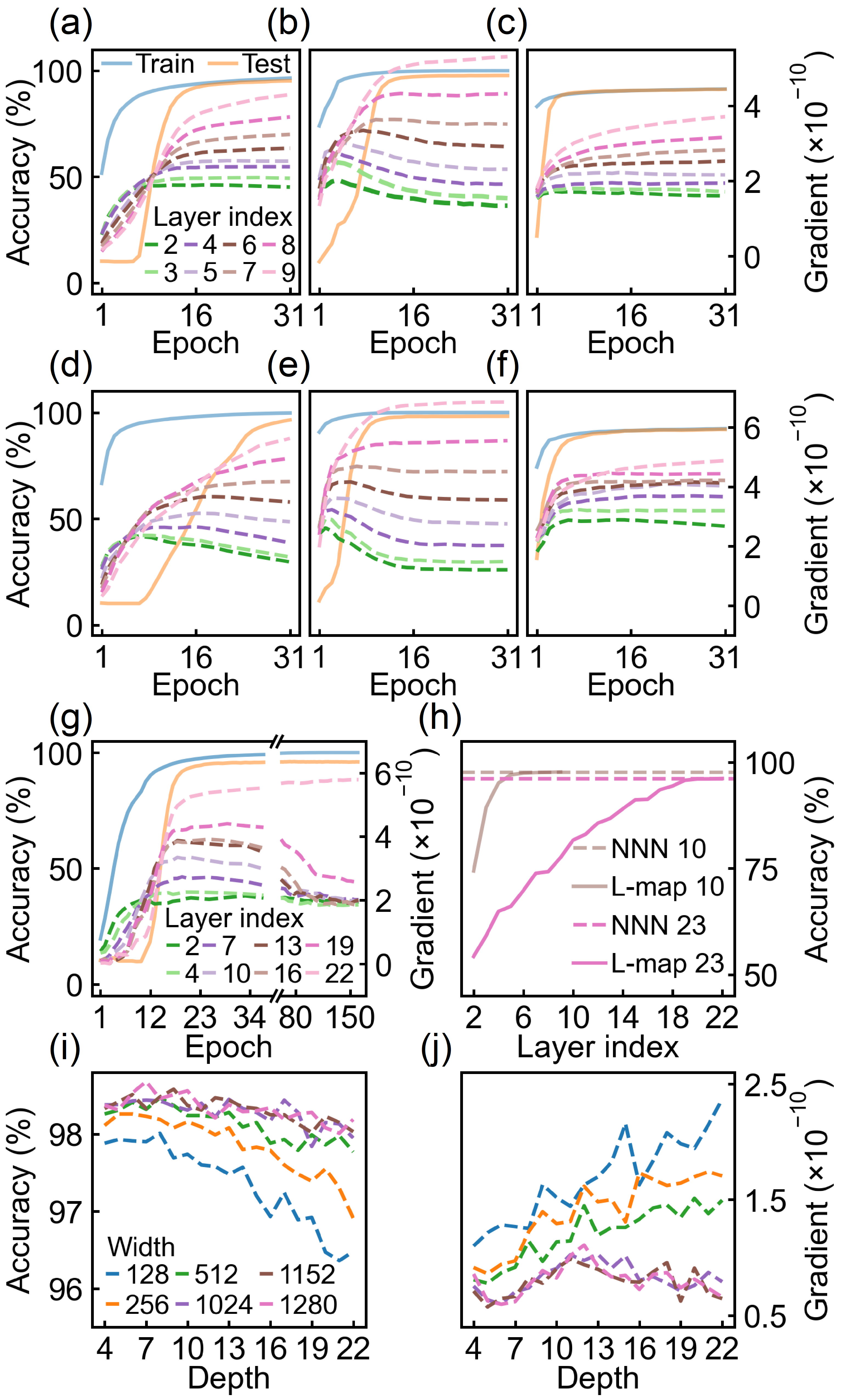}
\caption{Training dynamics and RPD analysis.  
(a)--(c) Evolution of training and test accuracy, together with RPD gradients, in a 10-layer DNN (width 512) trained with SGD. 
Panels (a) and (b) show results with ReLU activation and learning rates of 0.03 and 0.37, respectively, while panel (c) shows results with linear activation and a learning rate of 0.03. 
(d)--(f) Corresponding results obtained using the Adam optimizer. Batch sizes are 60000 for (d) and (f), and 20000 for (e). 
(g) Results for a 23-layer DNN (width 128) trained with Adam. 
(h) L-map pruning accuracy as a function of the starting layer for 10-layer and 23-layer networks. Note that layer index 1 denotes the input layer, so the x-axis starts from index 2 (the first hidden layer).
(i) Test accuracy as a function of depth for various widths. 
(j) First-layer RPD gradient as a function of depth for different widths.}
\label{fig3}
\end{figure}

Figs.~\ref{fig3}(a)--(c) show the evolution of training accuracy, test accuracy, and the RPD gradient for each hidden layer over training steps in a 10-layer DNN with a width of 512, trained using SGD. Here, the RPD gradient refers to the absolute slope of the RPD curve. ReLU activation is used in the first and second plots, with learning rates of 0.03 and 0.37, respectively. The third plot corresponds to training with a linear activation function and a learning rate of 0.03. 

The distribution of RPD gradients elucidates the specific mechanism of information processing. An overall increase in RPD gradients during the early training stage reflects the directional alignment of weight vectors induced by the OPT mode, which drives the weight vector directions from an initially isotropic distribution to more specific orientations. In the case of Fig.~\ref{fig3}(a), we see a clear phase transition: at the early stage of training, the RPD gradient is higher in
the earlier layers and lower in the deeper layers (denoted as phase I), but this pattern later reverses (phase II). In the other two plots, the learning process maintains a phase II–like gradient distribution from the outset.

Phase II represents an ideal structure of weight vector distribution. A low RPD gradient of the first hidden layer enables information extraction from a broad range of weight vector directions. As sample vectors become increasingly linearly separable across deeper layers, the RPD gradients increase gradually. Particularly, the RPD gradient of the first hidden layer characterizes the information extraction from the sample set. Fig.~\ref{fig3}(b) shows a lower first-layer RPD gradient than Fig.~\ref{fig3}(a), and thus the former achieves a superior test accuracy of 97.89\% over the  97.14\% of the latter.

The activation of NPT neurons enhances information extraction. The final RPD gradient distributions in Figs.~\ref{fig3}(a) and (c) are almost identical, implying that the two models extract largely similar information in the OPT mode. However, the former achieves the test accuracy of 97.14\%, whereas the latter reaches only 92.37\%. This discrepancy demonstrates that NPT neurons acquire additional information features by constructing and optimizing neuron combinations based on the weight vector distribution established by OPT.

Figs.~\ref{fig3}(d)--(f) present the results for DNNs trained using the Adam optimizer. The network architecture and activation functions exactly match those used in Figs.~\ref{fig3}(a)--(c). A batch size of 60000 is used in the first and third plots, while the second uses a batch size of 20000. Qualitatively similar phenomena are observed. Figs.~\ref{fig3}(d)--(f) achieve classification accuracies of 98.26\%, 98.29\%, and 92.46\%, respectively, which indicates the general superiority of Adam over SGD for training nonlinear DNNs. Again, the superiority can be attributed to the lower first-layer RPD gradients. 

Deeper networks may exhibit more distinct learning phases. Fig.~\ref{fig3}(g) shows the results for a 23-layer DNN with a width of 128, trained using the Adam optimizer and a batch size of 30000. In addition to phases~I and II, we observe a third phase characterized by the convergence of RPD gradients across almost all layers, with the exception of the last few, to approximately the same value. These results suggest that the learning process may transition through multiple phases with distinct layer-wise RPD gradient distributions, depending on the network architecture, training strategy, and hyperparameter configuration.

The degree of nonlinearity across DNN layers can be estimated via L-map pruning. We assess this by computing the pruning accuracy, which involves replacing layers from the \(l\)-th hidden layer to the output layer with the corresponding L-map. Fig.~\ref{fig3}(h) shows accuracy as a function of the starting layer for 10-layer and 23-layer DNNs, evaluated at the epoch corresponding to peak generalization performance.

The pruning accuracy differs from that of the original DNN when pruning begins at earlier layers, indicating the presence of NPT effects and inherent nonlinearity in those layers. As the starting layer moves deeper, the pruning accuracy increases, suggesting increasing linearity with depth. Beyond a critical layer, the accuracy stabilizes and becomes comparable with that of the original DNN, implying that the later layers are functionally equivalent to a linear perceptron and can be safely pruned. This behavior is desirable for reducing redundancy and conserving computational resources~\cite{NIPS1989_6c9882bb, 9795013, he2024matterstransformersattentionneeded, gromov2025the, men2024shortgptlayerslargelanguage, dalvi-etal-2020-analyzing, yom-din-etal-2024-jump}.

We then extend the RPD analysis to examine the effects of both network depth and width. Fig.~\ref{fig3}(i) presents test accuracy as a function of depth for DNNs with widths ranging from 128 to 1280. For a fixed width, accuracy initially increases with depth and then declines, with the reduction being more pronounced in narrower networks and more gradual in wider ones. The maximum accuracy is attained by networks with approximately 6 to 8 layers, a result that appears largely independent of width. For a fixed depth, accuracy generally increases with width and eventually saturates.

Since the RPD gradient in the first hidden layer plays a pivotal role in maximizing information extraction, we show the gradient across DNNs with varying depths and widths in Fig.~\ref{fig3}(j), evaluated at the epoch corresponding to the highest test accuracy, consistent with Fig.~\ref{fig3}(i). The key findings are as follows. First, as the width increases, the RPD gradient decreases. This trend correlates with the increase in test accuracy, suggesting that a higher proportion of NPT neurons in wider networks facilitates information extraction. Beyond a certain width, both the gradient and accuracy saturate. Second, the depth dependence of the RPD gradient reveals a pronounced minimum around 6 layers, indicating that network depth serves as a critical degree of freedom for minimizing the first-layer gradient and thereby maximizing information extraction from the training set.

An intriguing observation is that grokking consistently appears in the scenarios shown in Figs.~\ref{fig3}(a), (d), and (g), where the test accuracy remains at the level of random guessing even after training accuracy becomes high, before suddenly improving. Notably, the presence of grokking does not necessarily imply higher test accuracy. The underlying mechanism of grokking will be discussed in Section~\ref{sec:5} in terms of phase transitions.

\subsection{Attraction basin analysis}
\begin{figure}[ht]
\centering
\includegraphics[width=1\linewidth]{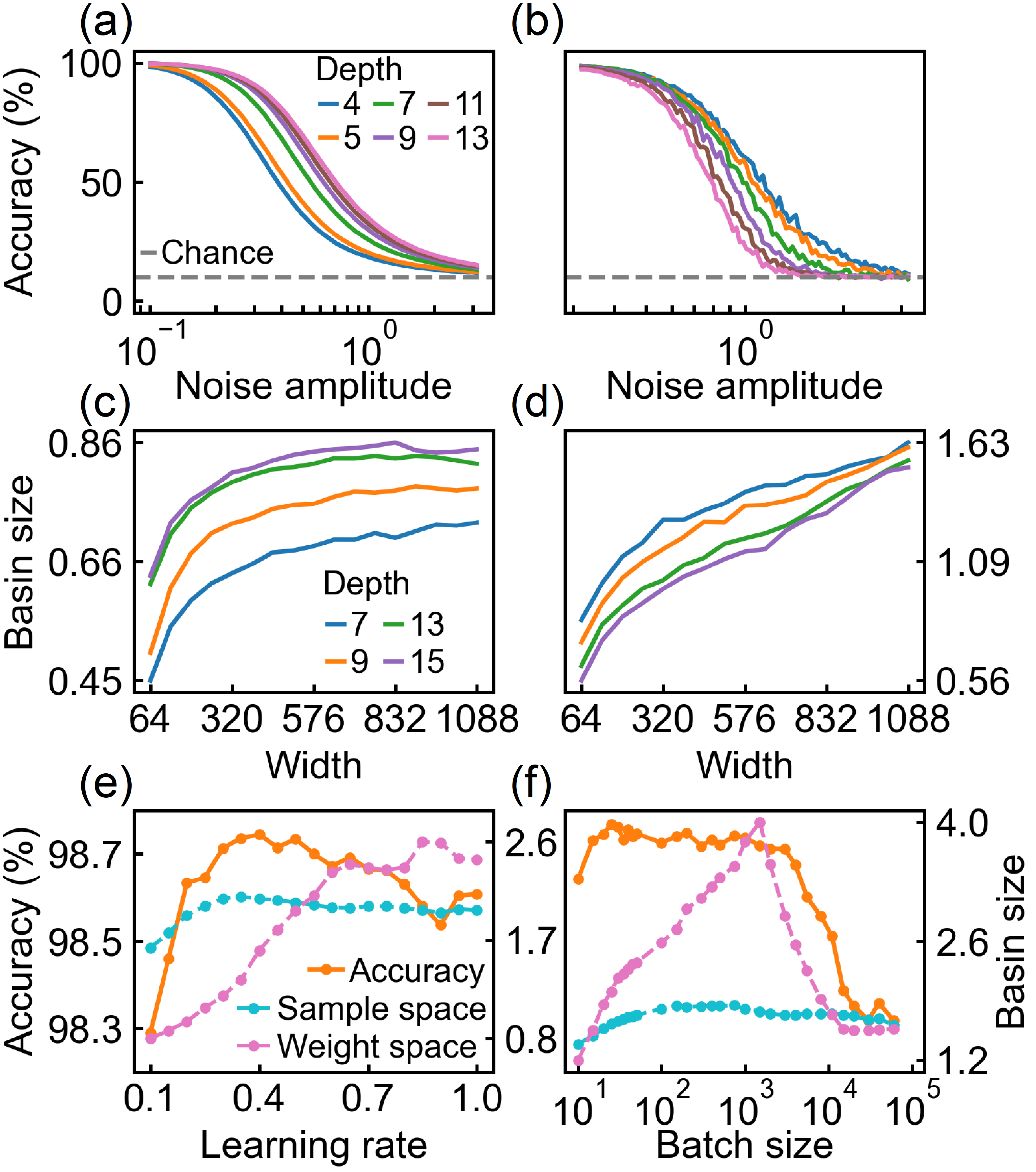}
\caption{Attraction basin analysis. 
(a) Accuracy of noisy training samples vs. noise amplitude. 
(b) Accuracy of training samples vs. noise amplitude under weight perturbations. 
(c), (d) Average attraction basin sizes in the sample and the weight space, respectively, as a function of network width for various depths. 
(e) Basin sizes in both the sample and weight spaces as a function of learning rate. 
(f) Basin sizes in both the sample and weight spaces as a function of batch size. 
In (e) and (f), sample-space basin sizes are scaled by a factor of 2 for clarity.}
\label{fig4}
\end{figure}

Fig.~\ref{fig4}(a) shows that increasing network depth enlarges the average attraction basin in the sample space. This highlights a key advantage of depth: the layer-by-layer iteration progressively expands the attraction basin in the sample space, thereby improving generalization ability. However, as shown in Fig.~\ref{fig3}(i), the optimal test accuracy occurs at a moderate depth, suggesting that deeper is not always better and implying that the attraction basin in the sample space does not have a monotonic relationship with test accuracy. The reason for this result is revealed in Fig.~\ref{fig4}(b), which shows that increasing depth reduces the attraction basin in the weight space, indicating that deeper networks negatively affect structural stability. Although test accuracy is typically insensitive to the size of the weight-space attraction basin, if the latter becomes too small, it can negatively impact test accuracy. The balance between these two attraction basins leads to the existence of an optimal depth for DNNs.

Figs.~\ref{fig4}(c) and (d) show the relationship between the average basin sizes in the sample and weight spaces as a function of width for DNNs with varying depths. For a fixed width, we observe that the sample-space basin size increases with depth, while the weight-space basin shrinks, consistent with the trends shown in Figs.~\ref{fig4}(a) and (b). For a fixed depth, both types of basins expand with width. It is important to note that in these calculations, we adopted a commonly used weight initialization strategy in DNN design, namely the Kaiming initialization, which samples the initial weights from 
$U\!\left(-\tfrac{1}{\sqrt{\mathrm{fan\_in}}},\, \tfrac{1}{\sqrt{\mathrm{fan\_in}}}\right)$, where $\mathrm{fan\_in}$ is the number of input connections~\cite{10.1109/ICCV.2015.123}. This strategy causes the initial weight bounds to shrink as the width increases. As seen in Fig.~\ref{fig4}(d), this results in a rapid increase in the attraction basin in the weight space with increasing width, which is the reason why test accuracy (Fig.~\ref{fig3}(i)) continues to improve as width increases. Without this initialization scheme, increasing the width could lead to a decrease in the weight-space attraction basin, thereby hindering the improvement in test accuracy. Based on this mechanism, more optimal initialization strategies exist, as discussed in the Supporting Information.

The attraction basin in the weight space maintains the structural stability of the network, and as long as it is not too small to cause structural instability, the dependence of test accuracy on it remains weak. This becomes especially evident when tuning hyperparameters in pursuit of ultimate accuracy. Fig.~\ref{fig4}(e) (with SGD) shows that increasing the learning rate leads to a monotonic expansion of the attraction basin in the weight space within the examined range. However, in this case, the maximum test accuracy essentially coincides with the largest attraction basin in the sample space. This suggests that, within this learning rate range, structural stability is maintained, and therefore, the attraction basin in the weight space does not significantly affect the network's accuracy. Fig.~\ref{fig4}(f) (with Adam) shows that the peak accuracy appears during the stage when both attraction basins expand simultaneously. 
This indicates that the expansion of attraction basins is a necessary condition for performance improvement, whereas such expansion alone is not sufficient to guarantee higher accuracy.
The rapid growth of the attraction basin in the weight space at smaller batch sizes may help sustain accuracy at the plateau, although it does not noticeably improve it. 
However, once the attraction basin in the weight space drops sharply, test accuracy also decreases rapidly, even though the attraction basin in the sample space remains relatively large. 
This asymmetry highlights the critical role of the weight-space attraction basin in preserving test performance.

\begin{figure}[ht]
\centering
\includegraphics[width=1\linewidth]{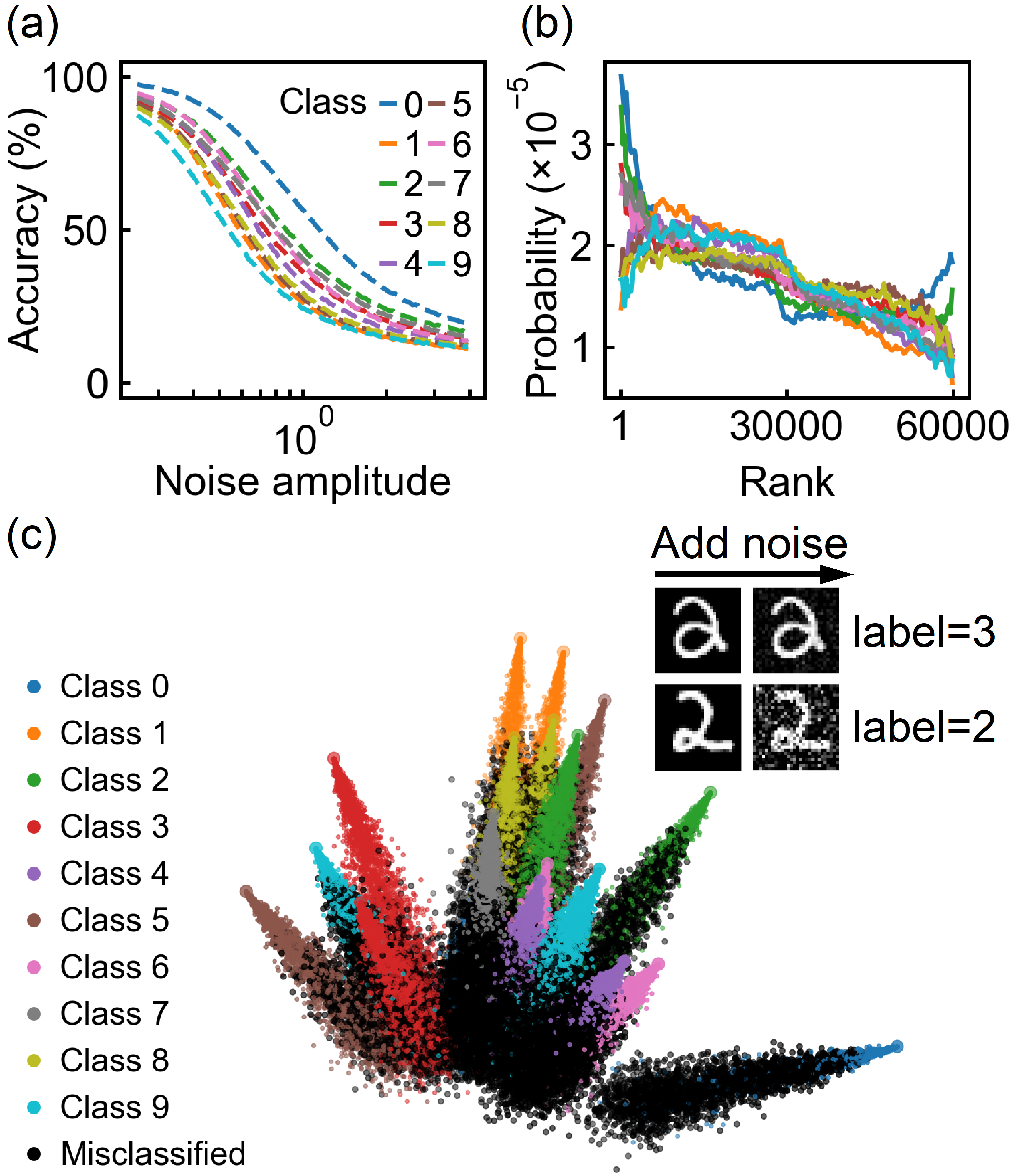}
\caption{Sample attraction basins and class-wise RPD curves for the 23-layer DNN in Fig.~\ref{fig3}(g). 
(a) Accuracy of noisy samples from ten classes versus noise amplitude. 
(b) Class-wise RPD curves at layer 2 (the first hidden layer). 
(c) Sample attraction basins in a 3D PCA projection of 20 digit samples. Samples with small attraction basins are more vulnerable to perturbations.} 
\label{fig5}
\end{figure}

The attraction basin analysis and RPD analysis can be further extended to different classes, offering finer-grained insights into DNN dynamics. Fig.~\ref{fig5}(a) shows the variation in classification accuracy across the ten classes (digits 0–9) as a function of noise amplitude. The results reveal significant differences in the attraction basins of different classes: for instance, digits 0 and 2 have relatively large attraction basins, while digit 9 has a much smaller one. By further examining the RPDs of different classes---illustrated in Fig.~\ref{fig5}(b), which depicts the first hidden-layer RPD after training across all ten classes---we find that digits 0 and 2 rely more heavily on OPT modes for information extraction and transformation, as their RPDs exhibit significantly higher density in the high-ranking (left-hand side) region. In contrast, digit 9 excites a greater number of NPT modes for its information processing, as the RPD distribution in this region is noticeably lower. It can be shown that digits 0 and 2 exhibit higher linear separability, while digit 9 shows lower linear separability (see Supporting Information). This suggests that the learning process handles different classes by utilizing different ratios of OPT and NPT neurons according to the specific characteristics of the sample set.

At the sample level, Fig.~\ref{fig5}(c) shows the attraction basins of 20 samples in a three-dimensional PCA projection, consisting of two samples from each of the ten digit classes. Most samples exhibit distinct effective attraction basins that remain non-contiguous, even among samples belonging to the same class. We also observe that the bases of the conical attraction basins tend to converge toward certain common regions, indicating that under sufficiently strong perturbations, all samples effectively behave as random patterns. More importantly, as one moves away from the apex of a conical basin, the number of perturbed samples attracted to other classes increases, giving rise to a fractal-like intermixing structure. This characteristic provides a theoretical foundation for the construction of adversarial examples. For instance, the sample of digit 2 located in the upper-right corner possesses a smaller attraction basin than another sample situated in the central region, implying that the former is far more susceptible to adversarial attacks. The inset illustrates a representative case: the former is misclassified as digit 3 under slight perturbations, whereas the latter can withstand even considerably stronger perturbations.

\section{phase transition and grokking}\label{sec:5}
RPD and attraction basin analyses provide a systematic framework for understanding neural network behavior, offering insights into the underlying learning dynamics. In this section, we investigate the mechanisms behind the well-known grokking phenomenon. Grokking is often described as a delayed transition from memorization to generalization after extended training \cite{liu2023omnigrok}. While this interpretation captures the behavior in shallow networks, we show that it does not accurately explain the dynamics in DNNs. Although multiple factors may contribute to grokking, a key prerequisite is a sharp shift in the phase of learning dynamics.

\subsection{Grokking in DNNs}
We demonstrate that the grokking effect observed in DNNs is triggered by the phase transition in conjunction with the training strategy involving BN (see \cref{eq:eq3}). To visualize this effect, we project the sample representations from the hidden layer adjacent to the output layer into two dimensions. Specifically, we insert a bottleneck layer with two neurons before the output layer \cite{PhysRevLett.132.057301}. Rather than retraining the model, we perform singular value decomposition on the weight matrix of the original final layer and extract the two leading right-singular vectors as principal directions. These directions are then used to construct the weights of the bottleneck layer. The resulting local fields \( h_1 \) and \( h_2 \) of the two bottleneck neurons are used as the coordinates in the two-dimensional representation.

\begin{figure}[ht]
\centering
\includegraphics[width=1.0\linewidth]{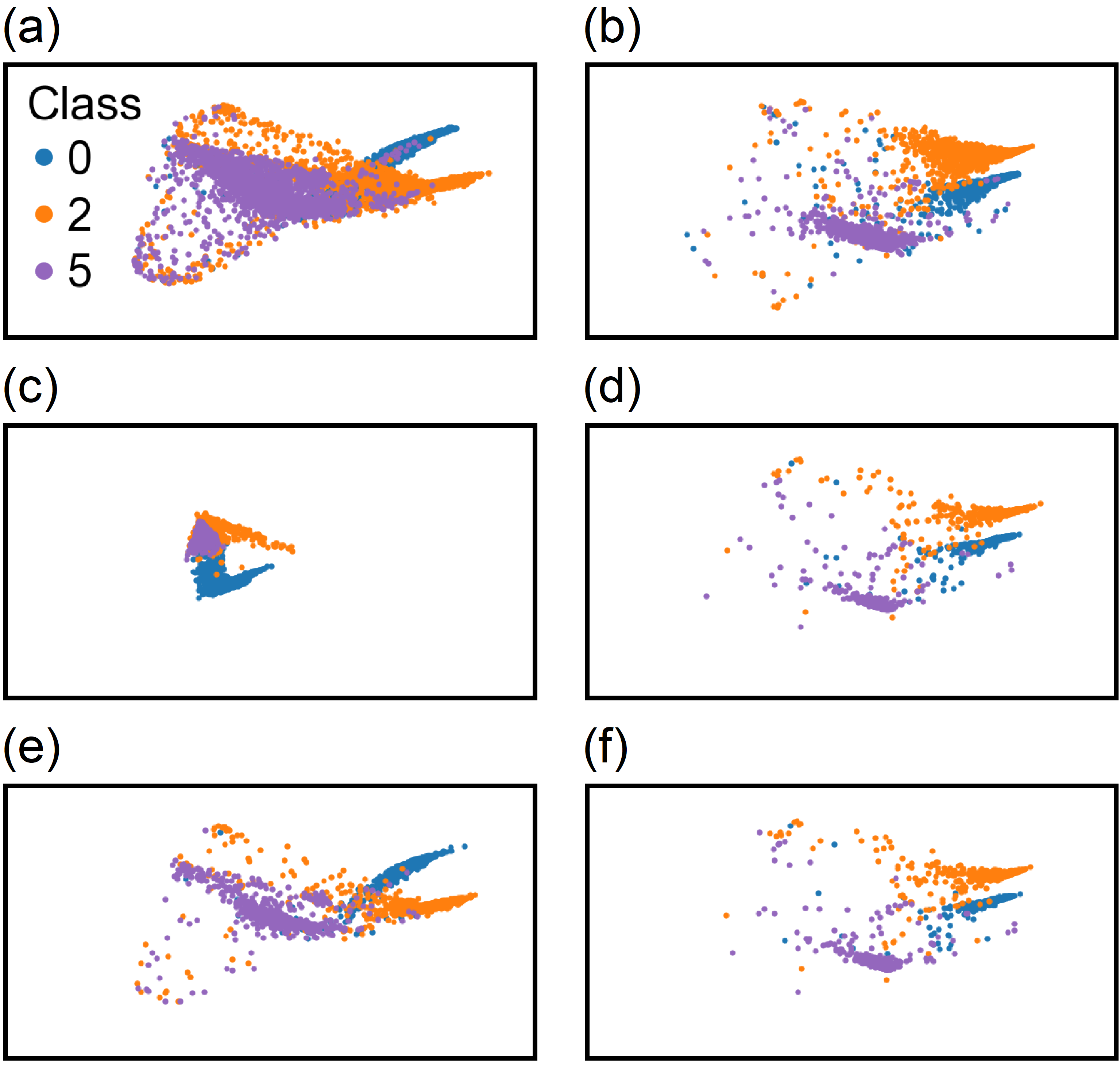}
\caption{Grokking mechanism in the 23-layer DNN. 2D representations of digits 0, 2, and 5 are shown before (left column) and after grokking (right column). 
(a), (b) correspond to the training set in training mode. 
(c), (d) show the test set in evaluation mode, while (e) and (f) show the test set in training mode. 
Note that a shift of test sample projections into the training attraction basins is observed only in evaluation mode after grokking.}
\label{fig6}
\end{figure}

For the 23-layer DNN, we visualize the 2D projections of training and test samples before and after grokking (Figs.~\ref{fig6}(a) and (c) vs. (b) and (d)). Prior to grokking, the test samples are projected outside the region occupied by the training samples. After grokking, the projection regions of the test and training samples overlap substantially, indicating that the test samples now fall within the attraction basins established by the training data. This overlap accounts for the abrupt rise in test accuracy.

This grokking scenario---where test sample projections collectively shift from outside into the attraction basins of training samples---can be attributed to the perturbation introduced by BN in evaluation mode, under a condition of phase transition in RPD gradient distribution. As described by \cref{eq:eq3}, BN relies on running estimates of mean and variance accumulated during earlier training stages. Phase transitions in learning dynamics (see Fig.~\ref{fig3}(g)) can significantly cause these historical estimates to diverge from the current true statistics, introducing a systematic bias in the inference for test samples. As training continues and the network settles into a stable learning phase, this discrepancy gradually diminishes. Consequently, test samples increasingly fall within the attraction basins of training samples, giving rise to the grokking phenomenon.

Indeed, when the training mode is applied at test time---by setting \(\alpha = 1\) in \cref{eq:eq3}---the grokking effect disappears. Under this setting, the projected regions of test samples (Fig.~\ref{fig6}(e)) already overlap with those of the training samples (Fig.~\ref{fig6}(a)) before grokking occurs. After grokking, the test sample projections (Fig.~\ref{fig6}(f)) remain similar to those in Fig.~\ref{fig6}(d) and coincide with the training sample projections (Fig.~\ref{fig6}(b)). In this case, maintaining consistent attraction basins for both test and training samples eliminates the conditions necessary for grokking to emerge.

Nonetheless, evaluation mode alone does not necessarily induce grokking. In cases where no phase transition occurs (Figs.~\ref{fig3}(b), (c), (e), and (f)), no notable grokking behavior is observed. This is because BN does not introduce significant distributional shifts within the same learning phase. Indeed, the 2D projection regions of test and training samples remain largely overlapping in these cases (see Supporting Information for details). These results underscore that grokking requires a transition in learning dynamics.

These findings suggest that the prevalent grokking behavior in deep neural networks cannot be fully accounted for by the conventional ``memorization-to-generalization'' narrative. Instead, it arises from phase transitions that cause BN to introduce statistical mismatches between training and test data.

\subsection{Grokking in shallow networks}
Under standard design settings, grokking behavior is generally not observed in shallow networks. However, as demonstrated in ref.~\cite{liu2023omnigrok}, grokking can emerge when specific strategies are employed. In their setup, a depth-4 \(784\text{-}200\text{-}200\text{-}10\) network was trained on 1000 samples with large initial weights and relatively small weight decay. We show that these conditions indeed give rise to pronounced grokking behavior.

\begin{figure}[ht]
\centering
\includegraphics[width=1.0\linewidth]{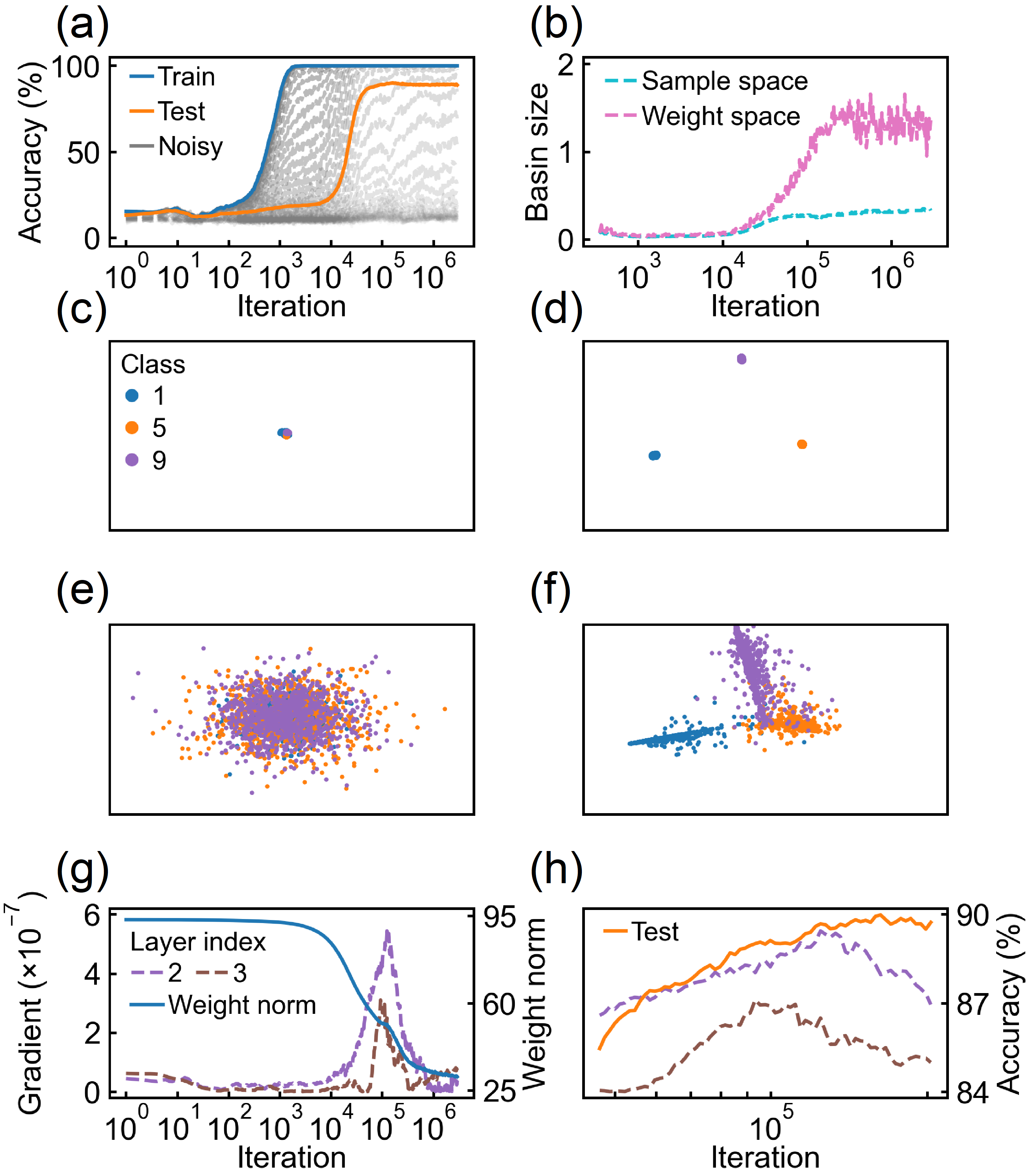}
\caption{Grokking mechanism in a shallow network. 
(a) Accuracy trajectories for training and test samples, including samples perturbed with different noise amplitudes. 
(b) Evolution of attraction basin sizes in both the sample and weight spaces, with a clear expansion coinciding with the onset of grokking. 
(c) and (d) show 2D projections of training samples for digits 1, 5, and 9 before and after grokking, respectively. 
(e) and (f) present the corresponding projections for test samples of the same digits before and after grokking. 
(g) RPD gradients of the two hidden layers, as well as the weight norm as a function of training steps. 
(h) Test accuracy evolution. The peak in test accuracy coincides with the decline in RPD gradients, followed by a gradual decrease upon further training. The RPD gradients are overlaid as dashed lines for reference (arbitrary scale).}
\label{fig7}
\end{figure}

In Fig.~\ref{fig7}(a), we reproduce the grokking phenomenon, characterized by a significant delay in test accuracy relative to training accuracy. The plot also shows the accuracy trajectories for samples with different noise levels. We see that when the noise amplitude remains below a certain threshold, the accuracy of noisy samples eventually approaches 100\%, indicating the formation of stable attraction basins. In contrast, when the noise amplitude exceeds this threshold, accuracy declines, suggesting that the samples fall outside the effective attraction basins.

The evolution of both types of attraction basins is shown in Fig.~\ref{fig7}(b) (not shown in the initial phase, as the basins have not yet formed). It shows a clear transition---an abrupt increase in both types of attraction basins coinciding with the onset of grokking. Prior to grokking, attraction basins are already formed for all training samples, as the training accuracy has reached nearly 100\%. However, the average basin size remains on the order of $10^{-2}$, whereas the deviation between test and training samples is approximately $10^{-1}$. After grokking, the attraction basin size increases beyond $10^{-1}$.

Figs.~\ref{fig7}(c)--(f) visualize the 2D projections of training and test samples for digits 1, 5, and 9 before and after grokking. Prior to grokking, the training samples form tightly clustered groups, while the test samples appear scattered and overlapping across classes, suggesting they fall outside the attraction basins. After grokking, the test projections become well-aligned with those of the training samples, indicating that they have entered the attraction basins. Therefore, grokking in this context emerges as test samples transition from regions outside to those inside the attraction basins where the training samples converge. Interestingly, we note a potential connection to the frequency principle \cite{Zhou2024GrokkingFreq}, where this transition aligns with a shift from capturing high-frequency components (due to large initialization) to learning generalizable low-frequency features, a mechanism that deserves further study. These combined patterns reveal a dynamical shift reminiscent of asymmetric Hopfield networks \cite{PhysRevE.70.066137}, where a sharp transition from a chaotic phase to a memory phase occurs as the attraction basins expand \cite{6795618,PhysRevE.72.066111}.

The RPD analysis reveals deeper mechanisms underlying grokking and the associated transitions in learning dynamics. Fig.~\ref{fig7}(g) displays the evolution of RPD gradients across the two hidden layers. It reveals that, prior to grokking, the RPDs in both hidden layers exhibit consistently small gradients. This observation suggests that the attraction basins at this stage are formed predominantly via the NPT mode. The activation of this mode arises from the specific architectural and training choices in the four-layer model. In particular, to prominently elicit the grokking phenomenon, the model is initialized with large weights and trained with a small weight decay rate of L2 regularization \cite{liu2023omnigrok}. This small decay rate, together with the correspondingly small learning rate, prevents weight vector concentration and thus suppresses the activation of the OPT mode. Furthermore, the small learning rate contributes to the formation of narrow attraction basins with NPT modes. Consequently, prior to grokking, the network remains in an NPT-dominated learning phase.
It is worth emphasizing that an NPT-dominated phase does not uniquely determine the generalization performance of the network.

Fig.~\ref{fig7}(g) also shows that, around the grokking point, the RPD gradients peak simultaneously across both hidden layers. This marks the onset of a new learning phase characterized by a high density of OPT neurons in both layers. The reason is as follows. During the NPT-dominated phase, the amplitude of the weight vectors gradually decreases (see the blue line in Fig.~\ref{fig7}(g)). Just before grokking, the amplitude of the weight vectors rapidly shrinks to very small values, making it easier even with a small learning rate to induce significant shifts in the directions of the weight vectors, thereby activating a large population of OPT neurons. These neurons, in turn, lead to a rapid expansion of the attraction basins, ultimately triggering the grokking transition. 

Continued training leads to a further decrease in the RPD gradients over time, signaling a sustained increase in NPT neurons and a return to the NPT-dominated phase, similar to the phase III in Fig.~\ref{fig3}(g). 
Although both the pre-grokking and late-training regimes are NPT-dominated, they correspond to fundamentally different structural states of the attraction basins.
This likely occurs because, at this stage, NPT neurons become more effective for further reducing the loss. As a result, test accuracy improves (Fig.~\ref{fig7}(h)), since consistently small RPD gradients allow the network to explore broader directions in weight space for information extraction.

However, ongoing training expands the attraction basin in sample space while the attraction basin in the weight space exhibits increasingly severe fluctuations (see Fig.~\ref{fig7}(b)). The fluctuations gradually destabilize the model and induce a decrease in test accuracy (see Figs.~\ref{fig7}(a) and (h)). Therefore, test accuracy requires the support of attraction basins in both the sample space and the weight space; coordination between the two is essential for achieving optimal performance.

This observation implies that training should be halted before excessive activation of NPT neurons occurs, thereby providing a theoretical justification for the widely used early stopping strategy \cite{prechelt2002early,yao2007early,NIPS2000_059fdcd9}.

\section{Discussion and conclusions}\label{sec:6}
The fundamental transformation modes and attraction basins serve as sensitive indicators and practical tools for probing learning dynamics, providing an intuitive framework for analyzing machine learning systems. The distinct information extraction mechanisms of OPT and NPT neurons give rise to different collective behaviors of weight vectors, which in turn shape the distribution of OPT and NPT neurons across hidden layers and result in distinct learning phases. The ratio of OPT to NPT neurons characterizes the degree of nonlinearity in each hidden layer, thereby resolving prior ambiguities in defining network linearity versus nonlinearity at the neuron level.

The attraction basins in the sample and weight spaces serve as complementary metrics for characterizing DNN dynamics. Their coordinated variations, together with the distribution of OPT-to-NPT neuron ratios across hidden layers, determine network performance and clarify the roles of architectural parameters and training strategies. In parallel, RPD analysis provides layer-, class-, and neuron-level resolution, while attraction basin analysis captures learning dynamics across multiple scales---from overall averages to class-specific and sample-specific basins. Together, these approaches reveal the internal dynamics of DNNs throughout training, transforming the so-called ``black box'' into a transparent framework and offering principled guidance for optimizing deep learning systems.

Learning phases, determined by training steps, initialization conditions, hyperparameters, dataset size, activation functions, and training strategies, are closely correlated with network performance. In particular, phase transitions give rise to significant phenomena such as grokking. On the one hand, we reveal that phase transitions are a prerequisite for the occurrence of grokking; on the other hand, we clarify that the grokking typically observed in DNNs and shallow networks originates from distinct mechanisms. In DNNs, grokking arises from the evaluation mode of the BN strategy: due to the influence of the phase transition, the projection regions of test samples are initially displaced from the attraction basins of the training samples, but as the learning phase stabilizes, they are gradually driven into them. By contrast, grokking in shallow networks originates from abrupt changes in the size of the attraction basins of training samples, aligning with the conventional definition of grokking as a shift from memorization to generalization.

Although this work primarily emphasizes the mechanistic understanding of learning models, it also carries immediate practical implications. For example, weight initialization is a critical step in training, and finding an effective initialization is particularly important for the Transformer architectures used in large language models \cite{yao2025an,NEURIPS2024_19c145aa}. Through the mechanisms revealed by the two types of attraction basin analyses, we can estimate optimal initialization values and precisely refine the conventional Kaiming initialization (see Supporting Information). Another application arises in the context of grokking. We find that pursuing grokking is not the optimal path to high performance. In contrast, keeping the network in the second phase from the outset avoids grokking and yields optimal performance more efficiently. This insight allows us to quickly assess whether a DNN, for given hyperparameters, begins in this favorable phase without resorting to long training runs to verify test accuracy. Such an approach can substantially reduce the computational cost of hyperparameter optimization in large models.

Finally, we emphasize that this paper focuses on presenting the fundamental concepts and principles of our framework. In fact, this analytical approach can be directly applied to other DNN architectures and datasets, including transformer models, and to reveal more detailed learning dynamics. We hope to showcase these insights in future research.

\begin{acknowledgments}
This work was supported by the National Natural Science Foundation of China (Grant Nos. 12247106, 12575042, and 11975189). H.Z. sincerely appreciates the beneficial discussions and suggestions from Professors Pan Zhang, Jie Yan, and Jiao Wang, which greatly enriched the theoretical perspective of this work.
\end{acknowledgments}

\section*{Conflict of interest}
The authors declare that they have no conflict of interest.

\section*{Supporting Information}
The supporting information is available online at \url{http://phys.scichina.com} and \url{https://link.springer.com}. The supporting materials are published as submitted, without typesetting or editing. The responsibility for scientific accuracy and content remains entirely with the authors.

\appendix
\setcounter{algorithm}{0}
\renewcommand{\thealgorithm}{A\arabic{algorithm}}
\section*{Appendix}

In the following, we present the algorithms for computing the RPD and for analyzing attraction basins.

\subsection*{A1 Algorithm for RPD}
We introduced the concepts of RPD and the L-map. Below, we provide the detailed steps of the algorithm. We take a 3-layer network (one hidden layer, see Fig.~\ref{fig2} in the main text) as an example. The computation of RPD for this network is shown in the following pseudocode (see Algorithm~\ref{alg:rpd}).

\begin{algorithm}[ht]
\caption{RPD algorithm for a three-layer neural network (Example: MNIST Dataset)}
\label{alg:rpd}
\begin{algorithmic}[1]

\REQUIRE
A trained three-layer neural network; hidden-layer pre-activation local fields
$H \in \mathbb{R}^{s \times d}$; output-layer weight matrix
$W \in \mathbb{R}^{d \times 10}$

\ENSURE
Rank matrix $P \in \mathbb{R}^{s \times d}$ and the corresponding
rank probability distributions

\STATE Initialize $P \leftarrow 0_{s \times d}$

\FOR{$i \leftarrow 1$ \TO $s$}
    \STATE Obtain the ground-truth label $y_i$ for sample $i$
    \STATE Set $j \leftarrow y_i$
    \STATE Compute $w_j = \mathrm{sign}(W_{:,j}) \in \{-1,1\}^{d \times 1}$
    \STATE Compute $H' = H \odot w_j^{\mathrm{T}}$
    \FOR{$k \leftarrow 1$ \TO $d$}
        \STATE
        $P_{i,k} \leftarrow
        1 + \left|
        \left\{
        m \in \{1,\dots,s\} \mid H'_{m,k} > H'_{i,k}
        \right\}
        \right|$
    \ENDFOR
\ENDFOR

\STATE Compute the RPD from $P$

\end{algorithmic}
\end{algorithm}

For deeper networks, as noted in the main text, to compute the RPD for a given layer $l$, we replace the output-layer weight matrix in the pseudocode with the matrix
\[
\mathbf{W}_{\mathrm{L\text{-}map}}^{(l)}
= \mathbf{W}^{(L)} \cdot \mathbf{W}^{(L-1)} \cdots \mathbf{W}^{(l+1)} .
\]

$\mathbf{W}_{\mathrm{L\text{-}map}}^{(l)}$ generalizes the output weights $\mathbf{W}$ in Algorithm~\ref{alg:rpd}, while $\mathbf{H}^{(l)}$ corresponds to the hidden pre-activations $\mathbf{H}$.

\subsection*{A2 Algorithm for the attraction basins}

In this work, we introduce the concept of the attraction basin in the sample space of a neural network to characterize the model's classification stability when its inputs are perturbed by random noise. From a statistical perspective, for a single sample, the randomness of the noise implies that---at a fixed noise level---one must perform multiple independent perturbation trials and average the results to estimate the true classification accuracy.

However, when dealing with large-scale datasets, one can exploit the ensemble averaging effect: at each noise level, apply only one independent perturbation to each sample and compute the overall accuracy. In a statistical sense, this result is equivalent to averaging over multiple perturbations per sample. This equivalence relies on the law of large numbers and the assumption that the noise is independently and identically distributed (i.i.d.), and therefore holds when the sample size is sufficiently large. In other words, for a single-sample case, multiple perturbations are required to ensure stability of the estimate, whereas for the multi-sample case, one or a few perturbations suffice to approximate the attraction basin characterization obtained from repeated experiments (see Algorithm~\ref{alg:sample_basin}).

\begin{algorithm}[ht]
\caption{Attraction basin size in sample space}
\label{alg:sample_basin}
\begin{algorithmic}[1]

\REQUIRE Training set $\mathcal{D}_{\mathrm{train}}$; trained neural network model $\mathcal{N}$
\ENSURE Attraction basin size $\eta^\ast$ in sample space

\STATE Uniformly select $n$ noise levels within a predefined interval,
$\{\eta_1, \eta_2, \ldots, \eta_n\}$

\FOR{each noise level $\eta \in \{\eta_1, \eta_2, \ldots, \eta_n\}$}
    \STATE Normalize pixel values of $\mathcal{D}_{\mathrm{train}}$ by dividing by $255$ to scale them to $[0,1]$
    \FOR{each sample $x \in \mathcal{D}_{\mathrm{train}}$}
        \STATE Generate Gaussian random noise $\xi \sim \mathcal{N}(0, I)$ with the same shape as $x$
        \STATE Construct perturbed sample $x' \leftarrow \mathrm{clip}(x + \eta \cdot \xi, 0, 1)$
    \ENDFOR
    \STATE Feed perturbed dataset $\mathcal{D}'(\eta)$ into $\mathcal{N}$ and compute accuracy $\mathrm{Acc}(\eta)$
\ENDFOR

\STATE Obtain the curve $\mathrm{Acc}(\eta)$ as a function of $\eta$
\STATE Define attraction basin size $\eta^\ast \leftarrow \min \{\eta \mid \mathrm{Acc}(\eta) \leq 50\%\}$

\end{algorithmic}
\end{algorithm}

Similarly, in the case of computing the attraction basin in weight space, a single weight perturbation simultaneously affects all samples. Strictly speaking, at a fixed noise level, multiple independent weight perturbations should be performed, and the results averaged to estimate the model’s overall performance. However, when the dataset is sufficiently large, sample averaging can significantly reduce the variance introduced by a single perturbation. As a result, the overall accuracy obtained under a single perturbation can already serve as a good approximation to the average over multiple perturbations. Therefore, in a similar manner, one or a few perturbations can be used as a practical approximation for characterizing the attraction basin in weight space (see Algorithm~\ref{alg:weight_basin}).

\begin{algorithm}[ht]
\caption{Attraction basin size in weight space}
\label{alg:weight_basin}
\begin{algorithmic}[1]

\REQUIRE
Training set $\mathcal{D}_{\mathrm{train}}$; trained neural network
$\mathcal{N}$ with parameters $\boldsymbol{\theta}$

\ENSURE
Attraction basin size $\eta^\ast$ in weight space

\STATE Uniformly select $n$ noise levels within a predefined interval,
$\{\eta_1, \eta_2, \ldots, \eta_n\}$

\FOR{each noise level $\eta \in \{\eta_1, \eta_2, \ldots, \eta_n\}$}
    \STATE Initialize perturbed parameters
    $\boldsymbol{\theta}^{(\eta)} \leftarrow \boldsymbol{\theta}$
    \FOR{each weight tensor $W \in \boldsymbol{\theta}^{(\eta)}$}
        \STATE Compute mean and standard deviation:
        $\mu_W \leftarrow \mathrm{mean}(W)$,
        $\sigma_W \leftarrow \mathrm{std}(W)$
        \STATE Standardize:
        $\widehat{W} \leftarrow (W - \mu_W)/\sigma_W$
        \STATE Sample Gaussian noise:
        $\varepsilon \sim \mathcal{N}(0, I)$
        \STATE Perturb in standardized space:
        $\widetilde{W} \leftarrow \widehat{W} + \eta \cdot \varepsilon$
        \STATE Transform back:
        $W' \leftarrow \sigma_W \cdot \widetilde{W} + \mu_W$
        \STATE Update $W \leftarrow W'$ in
        $\boldsymbol{\theta}^{(\eta)}$
    \ENDFOR
    \STATE Evaluate perturbed model
    $\mathcal{N}(\boldsymbol{\theta}^{(\eta)})$
    on $\mathcal{D}_{\mathrm{train}}$
    and compute accuracy $\mathrm{Acc}(\eta)$
\ENDFOR

\STATE Obtain the curve $\mathrm{Acc}(\eta)$ as a function of $\eta$
\STATE Define attraction basin size
$\eta^\ast \leftarrow \min \{\eta \mid \mathrm{Acc}(\eta) \leq 50\%\}$

\end{algorithmic}
\end{algorithm}

\end{document}